\documentclass[a4paper]{extarticle}
\usepackage{multicol}
\usepackage[margin=.8in]{geometry}
\usepackage{float}
\usepackage{graphicx}
\usepackage{amsmath}
\usepackage{makecell}

\usepackage{array}
\usepackage{multirow}
\usepackage[leftcaption]{sidecap}
\usepackage[font=scriptsize]{caption}
\usepackage{gensymb}
\begin{document}
\title{\textbf{IR Motion Deblurring}
}

\author{Nisha Varghese*         \and
        Mahesh Mohan M. R. \and A. N. Rajagopalan 
}

\date{ee19d003@smail.iitm.ac.in, ee14d023@ee.iitm.ac.in, raju@ee.iitm.ac.in}

\twocolumn
\maketitle

\begin{abstract}
Camera gimbal systems are important in various air or water borne systems for applications such as navigation, target tracking, security and surveillance. A higher steering rate (rotation angle per second) of gimbal is preferable for real-time applications since a given field-of-view (FOV) can be revisited within a short period of time. However, due to relative motion between the
gimbal and scene during the exposure time, the captured video frames can suffer from motion blur. Since most of the post-capture applications require blur-free images, motion deblurring in real-time is
an important need. Even though there exist blind deblurring methods which aim to retrieve latent images from blurry
inputs, they are constrained
by very 
high- dimensional
optimization thus incurring large execution times. On the other hand, deep learning methods for motion deblurring, though fast, do not
generalize satisfactorily
to different
domains (e.g., air, water, 
etc). In this work, we address the
problem of real-time motion deblurring in infrared (IR) images captured by a gimbal-based system. We reveal how \emph{a priori} knowledge of the blur-kernel can be used in conjunction with non-blind deblurring methods to achieve real-time performance.
Importantly, our mathematical model can be leveraged to create large-scale datasets with realistic gimbal motion
blur. Such datasets which are a rarity can be a valuable asset for contemporary deep learning methods. We show that, in comparison to the state-of-the-art techniques in deblurring, our method is better suited for 
practical gimbal-based imaging systems.

\end{abstract}

\section{Introduction}
\label{intro}
Camera gimbal systems, used in navigation, target tracking, security and surveillance, work by controlling the trajectory of a camera-mounted on a gimbal in order to obtain an extended FOV. During exposure, due to the relative motion between the scene and the gimbal, the captured images are typically degraded by motion blur. A motion blurred frame is the result of aggregation  of  different  world-to-sensor  projections  of  the  scene,  over  the exposure interval, onto the image sensor \cite{RefMDB}\cite{Purna}. Motion blur degrades image quality which can in turn lead to low probability of threat detection and higher false alarm rate during tracking and surveillance. Since most of the applications require blur-free images, motion deblurring is a very important problem. Moreover, such systems warrant motion deblurring in real-time. Even though there exist different effective algorithms to retrieve latent images from blurry inputs, they are not suitable for real-time applications due to their higher execution time.

Traditional methods for motion deblurring mainly comprise of two stages \cite{RefMDB}. First,
a motion blur model is designed to relate the sharp image to the observed blurred image via camera motion parameters (i.e., blur kernel). Second, on inverting this model, the camera motion parameters and  corresponding sharp photograph are estimated. The problem of deblurring is an  ill-posed one and sharp image estimation typically  necessitates the use of natural image priors \cite{Ref10}\cite{Ref11}. Traditional  deblurring methods typically proceed by alternating minimization (AM) of camera motion and sharp image. Further, AM proceeds  in a scale-space manner to accommodate large blurs to keep the optimization dimension low \cite{RefMDB}. 

A number of traditional methods exist for blind motion deblurring. In \cite{Ref2}, a parameterized geometric model of the blurring process is proposed in which the blurred image is represented using the transformed pixel value, due to rotations, and the time it spends at each rotation angle. Even though this method accommodates space variant blur kernels, a high computational cost renders it ineffective for real-time applications. There are situations in which the blur kernel can be assumed to be space-invariant thus reducing the complexity. The work \cite{Ref6} assumes a convolutional model and proposes a sparsity prior to estimate the blur kernel. The latent image is then reconstructed using standard Richardson-Lucy (RL) deconvolution \cite{Ref3}\cite{reflucy} algorithm. The method in \cite{Ref7} is aimed at handling blurred images in the presence of outliers. It models the data fidelity term so that the outliers have little effect on kernel estimation. A natural image prior known as extreme channel prior is proposed in \cite{Ref8} by taking advantage of both dark and bright channel priors for kernel estimation. Deblurring is done by alternating minimization of camera motion and sharp image. The method of \cite{Ref9} introduces an image prior based on a higher-order Markov random fields model termed as super-Gaussian fields to address the problem of sparsity of the prior. A scale normalization technique is introduced in \cite{Ref11} based on the \(L^p\) norm. \cite{Ref10} proposes a simplified sparsity prior of local patch-wise minimal pixels. Since it only requires values for non-overlapping patches, it is much simpler than the dark-channel prior. However, all these methods incur heavy processing time.

Another class of methods use deep learning networks. They work by optimizing the weights of a neural network, to emulate the non-linear relationship between
blurred-sharp image pair, using a training dataset. Both the effects of camera motion and dynamic object motion can be handled by these networks \cite{deblurgan}.  As these methods typically require only a single pass over the network for deblurring (unlike the AM), their processing time is comparatively quite less \cite{RefDL1}\cite{RefDL2}\cite{RefDL3}\cite{RefDL4}. 

Typically, in surveillance applications, infrared (IR) imagery is most suitable since it has a 24\(\times\)7 observation option during any atmospheric or illumination conditions \cite{IR2}\cite{IR}. We consider a commonly used gimbal set-up which extends the FOV via yaw-based gimbal motion. Our
system
consists of
an IR camera which is mounted on a gimbal set-up. The gimbal undergoes fast to and fro yaw motion to surveil the extended-FOV. Further, the gimbal motion is stabilized such that the motion disturbances in gimbal over other rotation directions (i.e., pitch and roll) are negligible. The speed is selected such  that a given FOV is revisited as quickly as possible.  The gimbal is equipped with an inertial measurement unit (IMU), consisting of gyroscopes and accelerometers.   The  IMU contains the  information to calculate the angular velocity of the device that it is mounted on. 

In this work, we specifically address the problem of real-time motion deblurring in IR images captured by a gimbal based system. It is important to note that state-of-the-art deblurring methods are designed for an unconstrained system such as random hand-held motion. Consequently, the number of unknowns and hence the processing time for these traditional methods is unacceptably high as they do not leverage the peculiarities of gimbal-based systems where camera motion is usually constrained or is tractable. This limits their applicability to the problem on hand. Deep learning networks, on the other hand, need a realistic training dataset with a large number of blurred and corresponding deblurred image-pairs. This requirement is often difficult to meet practically, especially in the case of IR images. Further, one pertinent problem is regarding generalization, i.e., a network trained using a particular dataset struggles when confronted with unseen examples from a different domain. 

In contrast to the afore-mentioned methods, we propose an effective real-time motion deblurring approach for IR images captured using gimbal-based systems. We advocate \emph{a priori} estimation of the blur kernel followed by \emph{non-blind} deblurring for real-time performance. Our main contributions are as follows:
\begin{enumerate}
\item This work provides a mathematical framework to encapsulate gimbal-based motion blur via extensive evaluations with
real gimbal data.
\item We leverage our derived model to devise a cost-effective non-blind deblurring solution that is 
almost real-time.
\item We experimentally verify that our method outperforms competing methods by 
a significant margin.
\item  Our model can be used to create large-scale motion blur datasets for gimbal-based systems that can be harnessed by deep
learning methods.
\end{enumerate}

\section{Our Approach}
\label{Non blind}
\begin{figure}
\centering
\setlength{\tabcolsep}{2pt}
\begin{tabular}{cc}

\includegraphics[height=1.3 cm]{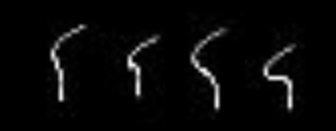} &
\includegraphics[height=1.3 cm]{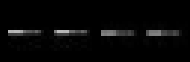}\\
(a)&(b)
\end{tabular}

\caption{PSF distribution for (a) real camera motion, and (b) yaw-only motion.}
\label{SV}       
\end{figure}

Blur due to camera rotations is typically associated with space-variant (SV) blur kernels \cite{Ref2}\cite{Kohler}\cite{Raj}. i.e., the nature of the kernel is different at different locations in the image. Figure \ref{SV}(a) shows plots of the point spread function (PSF) distribution in four corners of an image for real camera motion (using \cite{Kohler}). Even though SV methods \cite{Ref2}\cite{Kohler} are effective for real camera motion, their high computational cost renders them unsuitable for gimbal-based systems. The SV methods optimize for high-dimensional
camera motion, whereas the gimbal motion is predominantly a 1D rotational motion. The blur kernel corresponding to yaw-only motion is given in Fig. \ref{SV}(b). Note that PSFs produced by a general camera motion is quite different as compared to yaw-only motion PSFs. Observe that the PSF is mostly space-invariant. 
We will verify this interesting feature both theoretically and experimentally in the subsequent sections.

For space-invariant blurring, we can model the blurred image \(\mathbf{B}\) as a clean image \(\mathbf{L}\) 
convolved with a blur kernel \(\mathbf{k}\) and corrupted by noise \(\mathbf{n}\) \cite{Refb2}:
\begin{equation}
\mathbf{B}=\mathbf{L}*\mathbf{k}+\mathbf{n}
\label{Eq:conv_blur_model_ons}
\end{equation}
where $*$ denotes the convolution operator. Note that the noise \(\mathbf{n}\) captures the modelling errors, image noise and  the effect of outliers, e.g., saturated pixels and impulse noise, which cannot be well-modeled by the linear convolution model. Blind deblurring methods (mentioned in the previous section) estimate both \(\mathbf{L}\) and \(\mathbf{n}\). Patch-wise minimal pixel prior method \cite{Ref10} and normalized blind deconvolution \cite{Ref11} are some promising blind deblurring methods proposed recently, but lack real-time performance. 

The task of deblurring an image with a known blur kernel is known as \emph{non-blind} deblurring. Since the problem is only to find out the clean image \(\mathbf{L}\) from the blurred image \(\mathbf{B}\) and the blur kernel \(\mathbf{k}\), the processing time for non-blind deconvolution is comparatively far less than any blind deblurring method. In this work, we explore the exciting prospect of estimating the blur kernel a priori followed by non-blind deblurring.

Numerous non-blind deconvolution approaches exist, varying greatly in their speed and
sophistication. Wiener filter \cite{Refb2} is a traditional non-blind deconvolution method which finds the deblurred image from the knowledge of blurred observation and kernel along with mitigation of additive noise. The Richardson–Lucy (RL) algorithm \cite{Ref3}\cite{reflucy}, is an iterative procedure for recovering an underlying image that has been
blurred by a known PSF. The main problem in any fast non-blind deconvolution method is the presence of ringing artifacts in the deblurred results. The challenge addressed by RL algorithm
is in balancing the recovery of image details and suppression of ringing. There are works which have proposed new image priors to model natural image statistics so as to improve the quality of deblurred results, especially from a ringing perspective. \cite{Ref5} employs a hyper-Laplacian prior and the deconvolution is several orders of magnitude faster than existing
techniques. A patch-wise model is proposed in \cite{RefNBD1} that uses a Gaussian mixture prior. \cite{RefNBD2} tackles the problem of outliers to improve the quality of the deblurred results. 

Among the many existing non-blind deblurring methods, we zero-in on three methods, namely, Wiener filter (WF) \cite{Refb2}, RL deconvolution \cite{Ref3}\cite{reflucy} and deblurring using hyper-Laplacian priors \cite{Ref5} due to their attractive execution times. Our central idea is that if
we know the PSF for a blurred frame, then we can
speed-up deblurring by a large factor. For example, if we can compute accurately
the kernel for each steering rate in a cost-effective manner, then we can achieve considerable speed-up by employing one of the above three non-blind deblurring methods. However there is a caveat: Non-blind deblurring methods can yield good results only if the blur kernels are known accurately. We next present two blur-kernel estimation methods for the problem on hand.
\section{Blur kernel estimation using blur-sharp pair}
\label{BSpair3}
The strength of this method lies in the fact that it does not warrant knowledge of motion or camera parameters. Recollect that each pixel in a motion blurred image has contributions from the neighbouring pixels. It is well-known that temporal averaging of a number of consecutive sharp images, captured from a camera, yields a blurred image that resembles a motion blurred image. We make use of this fact to get the blurred images corresponding to different amounts of motion blur. For this purpose, the camera is panned very slowly (steering rate of 1 deg/sec is typical) so as to obtain blur-free images of the scene. Blurred images corresponding to different steering rates can then be obtained by suitably averaging a certain number of sharp frames. The blurred images along with their sharp image counterpart can be used to estimate the blur kernel corresponding to the motion blur content in the blurred image by solving Eq. (\ref{blur_sharp_psf}) using conjugate gradient.
\begin{equation}
\mathbf{k}=\arg\min_{\mathbf{k}}||\mathbf{B}-\mathbf{L}*\mathbf{k}||_2 \text{ \hspace{0.5cm}s.t.\hspace{0.3cm}} \sum \mathbf{k}=1
\label{blur_sharp_psf}
\end{equation}
where \(*\) denotes convolution operation, \(\mathbf{B}\) is the blurred image and \(\mathbf{L}\) is the corresponding sharp image. This method works for any type of blur kernel. We may take an initial guess for \(\mathbf{k}\) that is eventually refined to the original kernel after a pre-defined number of iterations.

\begin{figure}
\scriptsize
\centering
\setlength{\tabcolsep}{2pt}
\begin{tabular}{cc}
\centering
\includegraphics[height=2.5 cm]{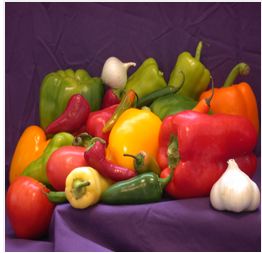}&
\includegraphics[height=2.5 cm]{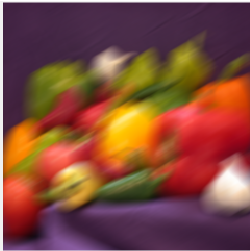}\\
(a) Sharp & (b) Blurred\\
\includegraphics[height=2.5 cm]{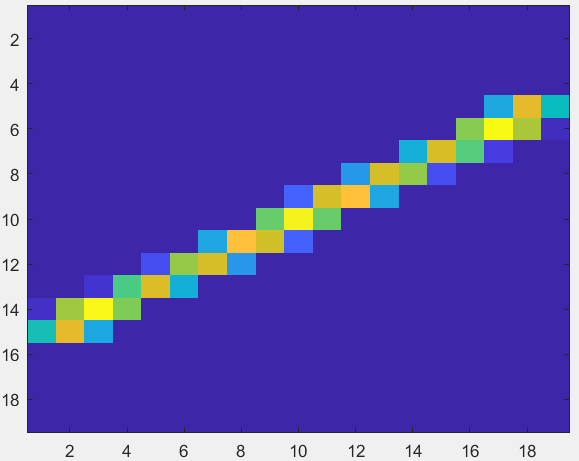}&
\includegraphics[height=2.5 cm]{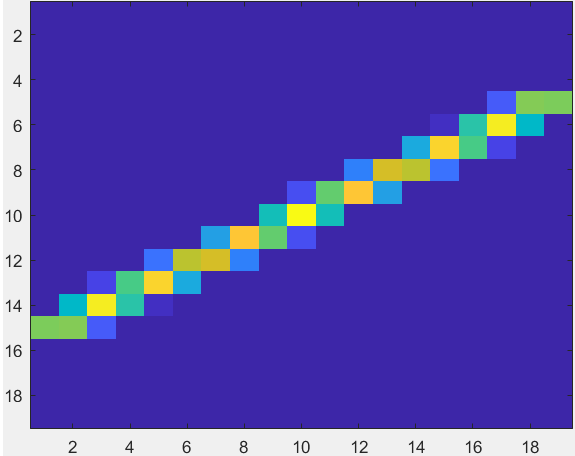}\\
(c) Estimated kernel & (d) Original kernel
\end{tabular}
\caption{Comparison of estimated kernel (c) (from blur sharp pair (a-b)), and the original kernel (d). Note that the Blur kernel estimation method using blur-sharp pair estimates PSFs similar to the ground truth.}
\label{im1}       
\end{figure}

Figure \ref{im1} gives an example of blur-sharp pair used for PSF estimation. The blurred image in Fig. \ref{im1}(b) is formed synthetically by blurring the sharp image (Fig. \ref{im1}(a)) with a diagonal motion blur kernel of size 21 pixels at an angle of 30\(\degree\) (Fig. \ref{im1}(d)). By solving Eq. (\ref{blur_sharp_psf}), the estimated kernel from the blur-sharp pair is given in Fig. \ref{im1}(c). It can be seen that the estimated blur kernel is quite close to that of Fig. \ref{im1}(d).

As mentioned earlier, frames captured with a steering rate of 1 deg/sec contain almost no motion blur as the motion is minimal. We treat these frames as sharp frames. A blurred image can be formed by averaging \(N\) images where \(N \in {1, 2, 3,\cdots, 13}\). Thus, we can arrive at the average image as the blurred image and the center frame as the sharp image. As an example, if \(N=5\), then we average 5 consecutive frames to get the blurred image \(\mathbf{B}\) and frame number 3 is treated as the sharp image \(\mathbf{L}\). If \(N\) is an even number, frame number \(\frac{N}{2}+1\) is taken as the sharp image. We can then find kernel \(\mathbf{k}\) from the blurred image \(\mathbf{B}\) and the sharp image \(\mathbf{L}\) using Eq. (\ref{blur_sharp_psf}). A uniform kernel of a certain size can be taken as the initial estimate of \(\mathbf{k}\). It is better to take planar scenes to form the average image to get a consistent kernel since depth differences in the scene can lead to errors in the estimate of \(\mathbf{k}\).

Thus we can get an estimate of the PSF (i.e., blur kernel) from blur-sharp pairs for different values of \(N\). Note that a large \(N\) increases the blur content in the averaged frame and hence the size of the blur kernel. The blur kernel corresponding to different values of \(N\) can be estimated to match the blur in images corresponding to different steering rates (from 10 deg/sec to 60 deg/sec). The value of \(N\) to be taken from steering 1 deg/sec to make the blur equivalent to the blur in a given steering rate \(s_r\) can be found out either empirically by matching the deblurred results for different steering rates or it can be derived if the exposure time \(t_{exp}\) of the camera is known. The rotation angle of a camera with steering rate \(s_r\) and exposure time \(t_{exp}\) is given by
\begin{equation}
    \theta_{s_r}=t_{exp}\cdot s_r
\end{equation}
A steering rate of 1 deg/sec covers 1\(\degree\) in 1 sec. If \(f_r\) is the frame rate (in frames/sec) then, the number of frames is given by
\begin{equation}
\label{SR}
    N=f_r\cdot \theta_{s_r}
\end{equation}

\section{Analytical modeling of Blur kernel}
\label{analy}
In this section, as an alternative, we wish to find the PSF mathematically for the case of gimbal motion that pans a scene with an IR mounted camera with known intrinsic camera parameters. Unlike the method discussed in Section 3, this is an entirely analytical approach to find the PSF from camera motion and camera intrinsics and does not need the blurred observations at all.

Consider a 2D impulse signal (a value 1 at the center of the image and 0 elsewhere). During exposure, due to relative motion between camera and the scene, the impulse signal will be distributed across adjacent pixels in the captured image. This forms the PSF for that motion model. To find the PSF analytically, we first assume a maximum pixel spread \(s\) (for each steering rate), from the center of the image, due to the camera motion. We next find the rotation angle of the camera corresponding to every pixel displacement upto the maximum pixel displacement \(s\). Then we find the transformation of the center pixel value using homography equations for every rotation angle of the camera. First, we will derive the equation for the rotation angle corresponding to a pixel spread that occurs in the image due to yaw-rotation induced motion blur.
\subsection{Rotation angle \(\theta\) corresponding to a pixel spread}
\begin{figure}
\centering
\includegraphics[height=5 cm]{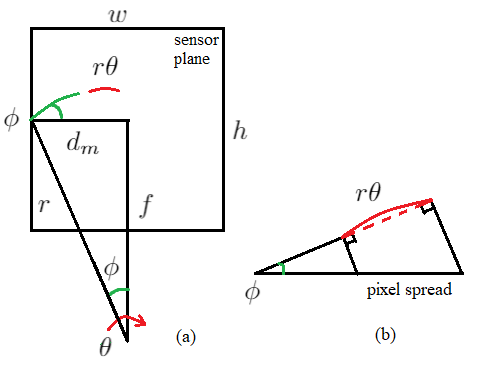}

\caption{Relation between rotation angle and the pixel spread.}
\label{fig2}       
\end{figure}
Figure \ref{fig2} illustrates the situation on hand. The camera is rotated by an angle \(\theta\) where \(w\) and \(h\) are the dimensions of the image, \(f\) is the focal length of the camera, \(d_m\) is the maximum displacement from the center of the image, and \(r\) is the distance from the edge pixel point to camera center. 

When the camera rotates by an angle \(\theta\), an arc with length \(r\theta\) will be formed on the rear of the sensor plane whose image is the pixel spread on the image plane. Assuming a linear arc length \(r\theta\), we can write the arc length as the projection from the sensor plane (as seen in Fig. \ref{fig2}(b)). Therefore,
\begin{equation}
r\theta = p_s\cdot\cos\phi
\label{eqn:2}
\end{equation}
where \(\phi\) is the angle between the arc and the sensor plane, and \(p_s\) is the corresponding pixel spread. From the figure, it can be seen that the angle between the arc and the sensor plane is equal to the angle between \(f\) and \(r\). Therefore \(\cos\phi\) can be written as \(\cos\phi=f/r\) and \(r\) can be written in terms of \(f\) and \(d_m\). Therefore Eq. (\ref{eqn:2}) becomes
\begin{equation}
  \sqrt{f^2+d_m^2}\theta = p_s \cdot \left(\frac{f}{ \sqrt{f^2+d_m^2}}\right)
  \end{equation}
  Hence, the rotation angle \(\theta\) corresponding to the pixel spread can be written as
  \begin{equation}
      \theta = p_s \cdot \left(\frac{f}{f^2+d_m^2} \right)
      \label{eqn4}
  \end{equation}

We next list the steps to find the blur kernel corresponding to the pan motion of the camera given the intrinsic and motion parameters of the camera.
\begin{enumerate}
    \item The intrinsic camera matrix is given by,
    \begin{equation}
        \mathbf{K}=\begin{bmatrix}
f & 0 & w/2\\
0 & f & h/2\\
0&0&1
\end{bmatrix}
    \end{equation}
    If the focal length (\(f\)) of the camera is not known, then it can be found out using FOV \(\alpha\) and the frame dimensions \(w\) and \(h\) as,
    \begin{equation}
    \label{f}
        f=\frac{\sqrt{h^2+w^2}}{2\tan(\alpha/2)}
    \end{equation}
    \item Find the rotation angles of the camera \(\theta_1\) and \(\theta_{max}\) corresponding to 1 pixel spread and maximum pixel spread \(s\) respectively, using Eq. (\ref{eqn4}).
    \item Since the rotation is only about the y axis, the rotation matrix \(\mathbf{R}\) corresponding to the rotation angle \(\theta\) can be written as
    \begin{equation}
        \mathbf{R}=\mathbf{R}_y=\begin{bmatrix}
        \cos\theta&0&\sin\theta\\0&1&0\\-\sin\theta&0&\cos\theta
        \end{bmatrix}
    \end{equation}
    Find the rotation matrices for the rotation angles from \(-\theta_{max} \text{\ to} +\theta_{max}\) with step size \(\theta_1\).
    \item For the pure rotation model, the homography \(\mathbf{H}\) \cite{Ref2} can be written as,
    \begin{equation}
        \mathbf{H}=\mathbf{KRK}^{-1}
    \end{equation}
    Find all the homography matrices corresponding to every rotation obtained in the previous step.
    \item Take the center pixel value \(\mathbf{x}\) of the image and find its transformed point \(\mathbf{x'}=\mathbf{H}\mathbf{x}\) for all \(\mathbf{H}\) and mark 1 at every \(\mathbf{x'}\) if \(\mathbf{x'}\) is an integer. If \(\mathbf{x'}\) is not an integer, then distribute 1 to the neighbouring positions of \(\mathbf{x'}\) using bilinear interpolation.
    \item Normalise the PSF so that it sums to 1.
    \item This will give the PSF corresponding to the center position.
    \item If we are interested in obtaining PSFs corresponding to different positions of the image, repeat steps 5 and 6 for pixel locations other than the center position. 
\end{enumerate}
The above derivation can be extended in a straightforward manner to general gimbal-trajectories.
\subsection{Selection of maximum pixel spread}
\label{analy21}
Note that in order to model blur kernel analytically, we assume that the maximum pixel spread \(s\) in the blur kernel is known for a particular steering rate. It equals half the kernel length corresponding to that steering rate. A higher steering rate will imply a higher value of \(s\). If the exposure time is known, then the maximum spread \(s\) can be determined a priori as follows.

Let the exposure time of the camera be \(t_{exp}\) and the steering rate be \(s_r\). The total angle of rotation \(\phi'\) during the exposure time can be written as 
\begin{equation}
    \phi'=s_r\cdot t_{exp}
\end{equation}
Then by using Eqn. (\ref{eqn4}), we can find the maximum pixel spread \(s\) from the center as 
\begin{equation}
\label{s}
    s=\left(\frac{\phi'}{2}\right)\cdot \left(\frac{f^2+d_m^2}{f} \right)
\end{equation}

\section{Experimental Results}
\noindent\textbf{Datasets:} We present results on real IR images corresponding to seven different steering rates and captured from an actual gimbal system that pans a high-altitude scene. The steering rates are 1 deg/sec and 10 to 60 deg/sec with a step-size of 10. Each frame is of size 558 \(\times\) 481 pixels. The video corresponding to steering rate of 1 deg/sec contains 3067 frames while the other 6 videos contain on the average about 675 frames. We perform deblurring on a machine with Intel Xeon e5-2630v4 processor \(@\) 2.2 GHz. 

In this section, we provide results of our method i.e., non-blind deblurring with WF, RL and \cite{Ref5} using the blur kernel estimated i) from blur-sharp pair, and ii) analytically. For comparisons, we examine different competing methods. Among blind deblurring methods, we choose \cite{Ref11} as it is the best in its class. We also use the PSF returned by \cite{Ref11} to perform non-blind deblurring with Wiener filter, RL \cite{Ref3}\cite{reflucy} and \cite{Ref5}.

\begin{figure}
\scriptsize
\centering 
\setlength{\tabcolsep}{2pt}
\begin{tabular}{ccc}
\includegraphics[height=2.2 cm]{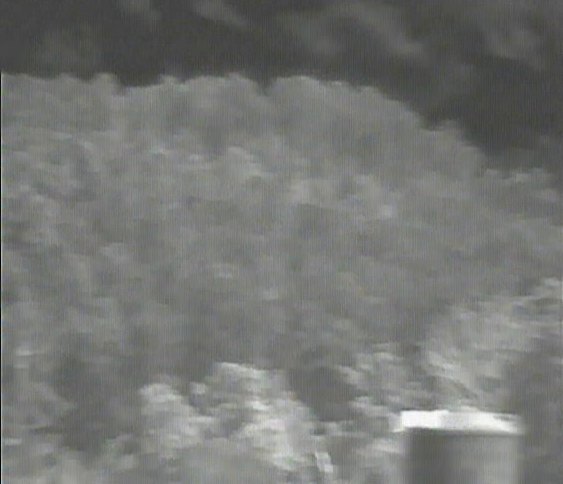} &
\includegraphics[height=2.2 cm]{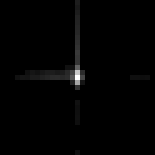} &
\includegraphics[height=2.2 cm]{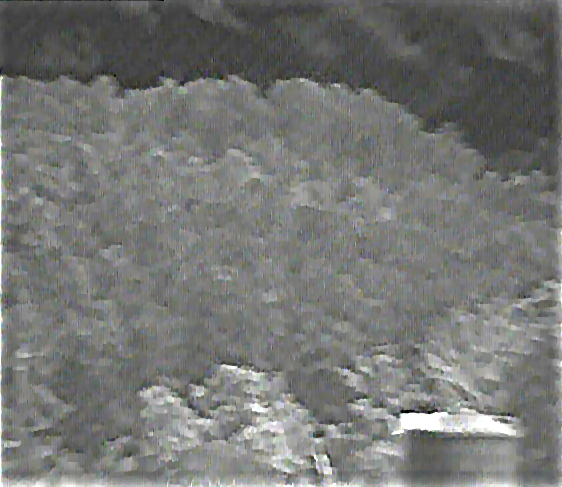} \\
(a) Blurred - SR50 & (b) Kernel & (c) Deblurred output\\
\includegraphics[height=2.2 cm]{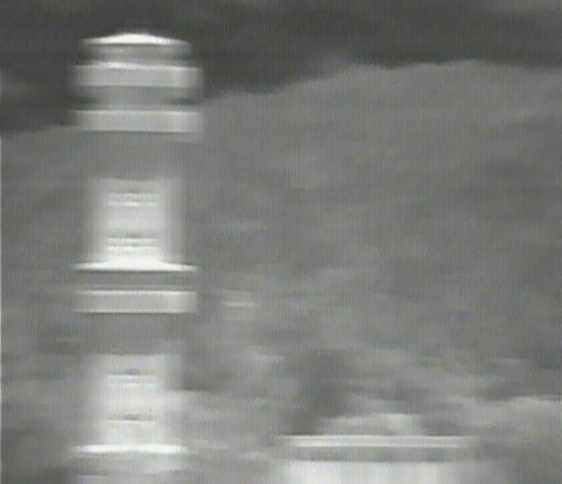} &
\includegraphics[height=2.2 cm]{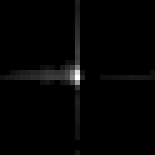} &
\includegraphics[height=2.2 cm]{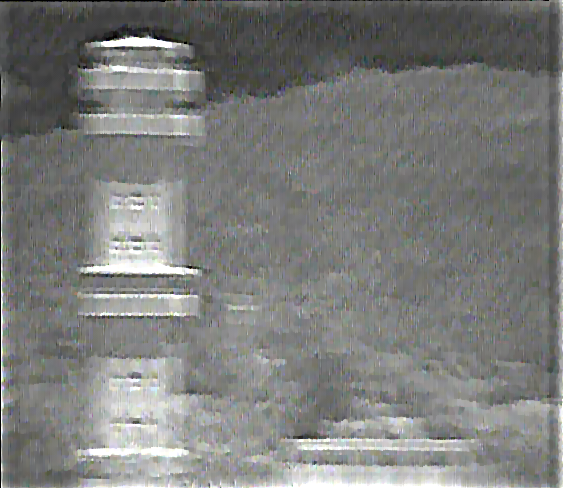} \\
(d) Blurred - SR60 & (e) Kernel & (f) Deblurred output\\
\end{tabular}
\caption{Deblurred results of blind deblurring method \cite{Ref11}. (a): steering  rate  50  deg/sec, (d): steering rate 60 deg/sec.} 
\label{NBD}
\end{figure}

We begin by presenting qualitative comparisons for competing methods. Deblurring results for steering rate 50 deg/sec and 60 deg/sec using \cite{Ref11} are given in Fig. \ref{NBD}. While the results are somewhat satisfactory, the processing time is about 10 minutes per image which is prohibitively high.

\begin{SCfigure*}
\scriptsize
\centering 
\setlength{\tabcolsep}{2pt}

\caption{Comparisons of non-blind deblurring methods using the kernel obtained from \cite{Ref11}. (a): steering rate 50 deg/sec, (e): steering rate 60 deg/sec.}
\begin{tabular}{cccc}
\includegraphics[height=2.5 cm]{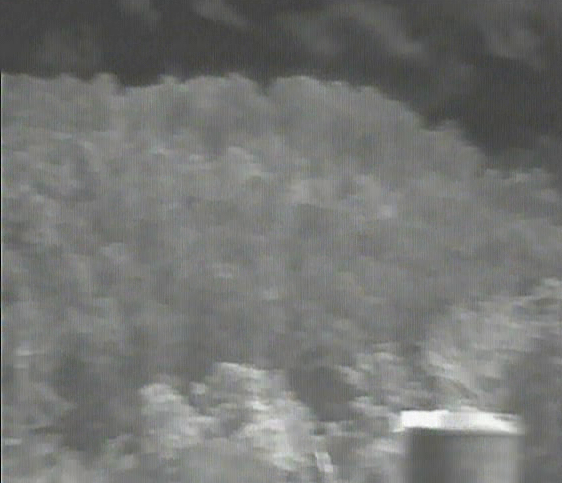}   &
\includegraphics[height=2.5 cm]{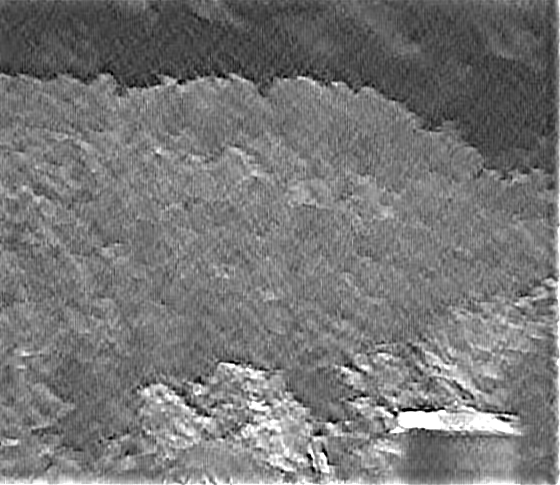} &
\includegraphics[height=2.5 cm]{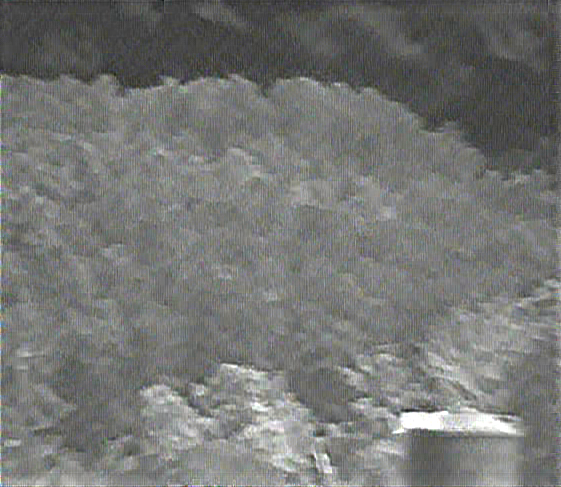} &
\includegraphics[height=2.5 cm]{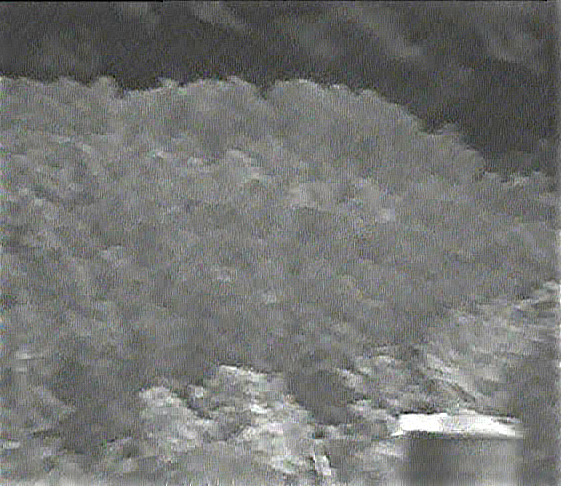}\\
(a) Blurred - SR50 &(b) Krishnan et al. \cite{Ref5}&(c) RL &(d) Wiener 
\\
\includegraphics[height=2.5 cm]{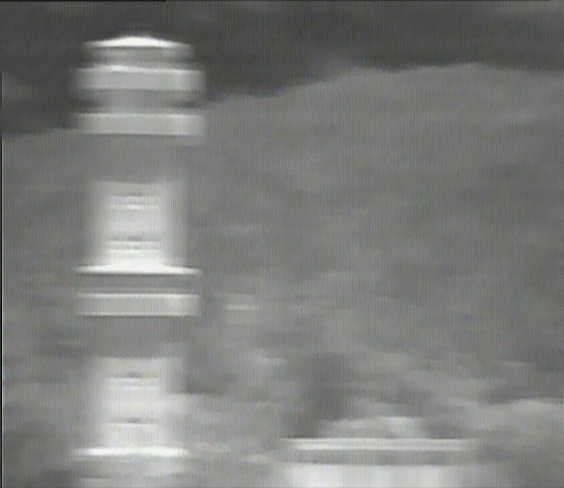}  &
\includegraphics[height=2.5 cm]{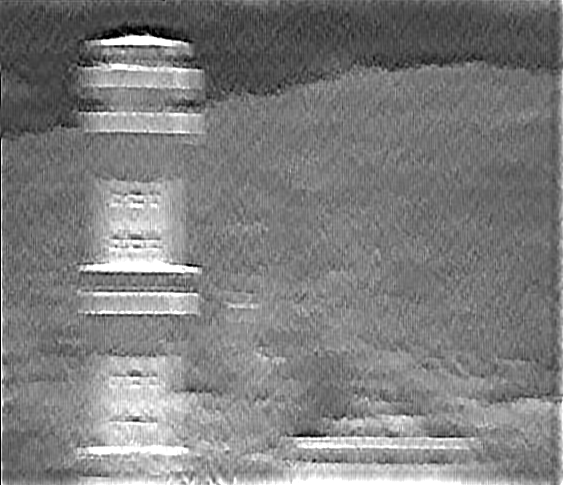} &
\includegraphics[height=2.5 cm]{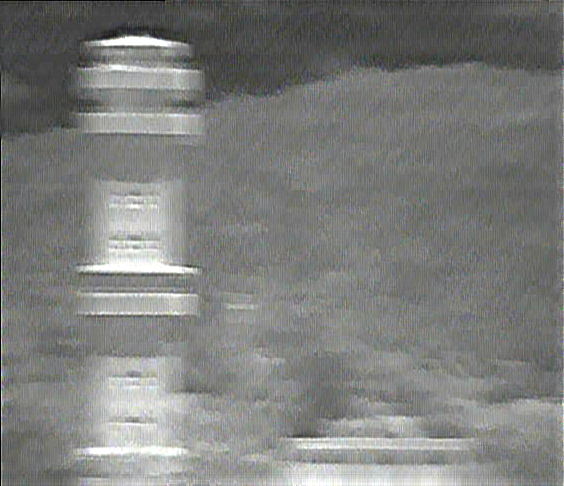} &
\includegraphics[height=2.5 cm]{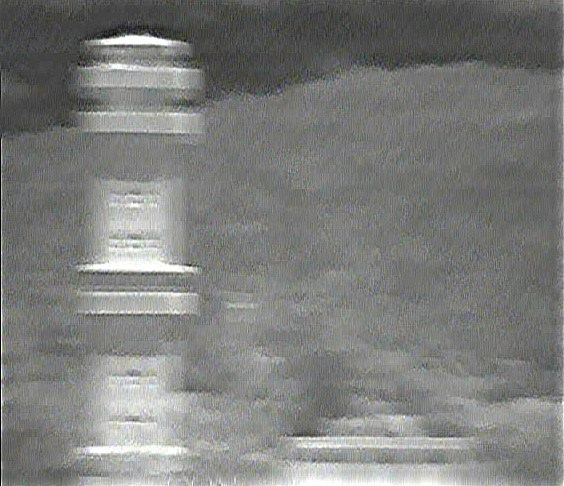}\\
(e) Blurred - SR60 &(f) Krishnan et al. \cite{Ref5}&(g) RL &(h) Wiener 
\end{tabular}

\label{NBD_kernel} 
\end{SCfigure*}
Since blind deblurring methods cannot be implemented in real-time, we next use three non blind deblurring methods (as explained in section \ref{Non blind}) with a known PSF. Since the blind method \cite{Ref11} gives good results, it stands to reason that it is able to fairly model the underlying blur kernel associated with the blurred images. Hence, we use the PSFs returned from \cite{Ref11} and perform non-blind deblurring using Wiener filter, RL \cite{Ref3}\cite{reflucy} and \cite{Ref5}.

\emph{Edge-tapering}: Before applying any non-blind deblurring method, we perform edge-tapering on the blurred images to remove boundary related ringing  artifacts that stem from high frequency drop-off during deblurring. Edge-tapering is done using a PSF \(\mathbf{k}_{et}\) (we use a Gaussian filter of size 30 pixels and \(\sigma=10\)), which gives an output image \(\mathbf{J}\) which is the weighted sum of the original
image \(\mathbf{I}\) and its blurred version. The weighting array is determined by the autocorrelation function of \(\mathbf{k}_{et}\). The image \(\mathbf{J}\) contains the original image portion, \(\mathbf{I}\), in its central region, and the blurred version of \(\mathbf{I}\) (using \(\mathbf{k}_{et}\)) near the edges. 

The results given in Fig. \ref{NBD_kernel} are comparable to that of \cite{Ref11} but come at a fraction of the computational cost incurred by \cite{Ref11}.

\subsection{Results using PSF from blur-sharp pair}
\label{bspair_result}

We generate blurred images by averaging a set of sharp frames. In our dataset, frames captured with a steering rate of 1 deg/sec can be treated as sharp frames since the motion blur is minimal for this steering rate.

\begin{figure}
\scriptsize
\centering 
\setlength{\tabcolsep}{1pt}
\begin{tabular}{ccc}
\includegraphics[height=1.9 cm]{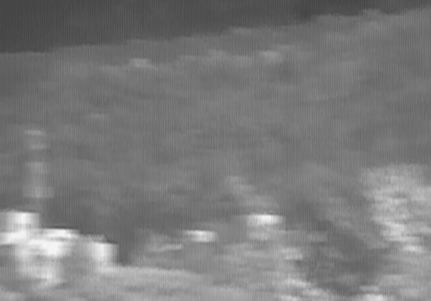} &
\includegraphics[height=1.9 cm]{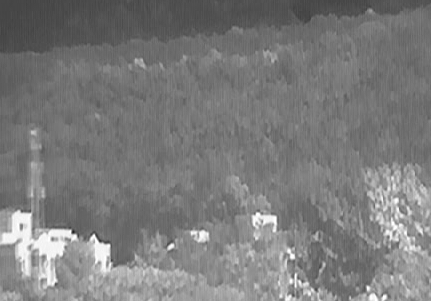} &
\includegraphics[height=1.9 cm]{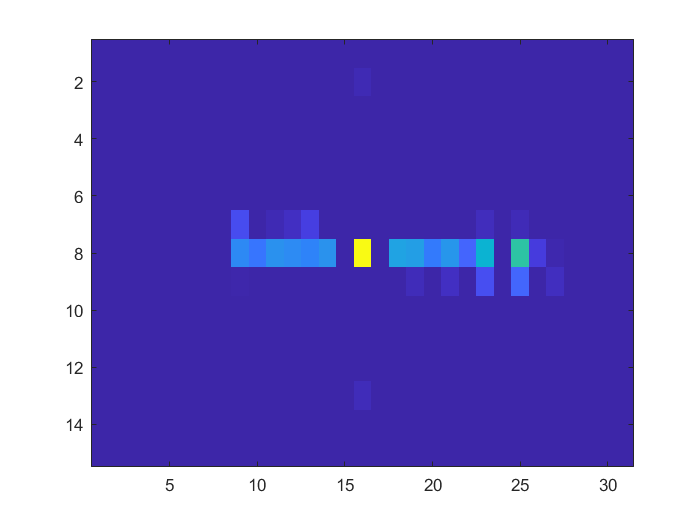} \\ 
(a) Mean of 6 frames &(b) Center-4th frame &(c) Kernel\\
\includegraphics[height=1.9 cm]{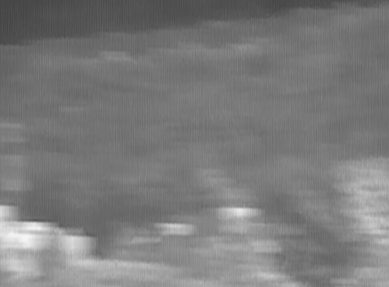} &
\includegraphics[height=1.9 cm]{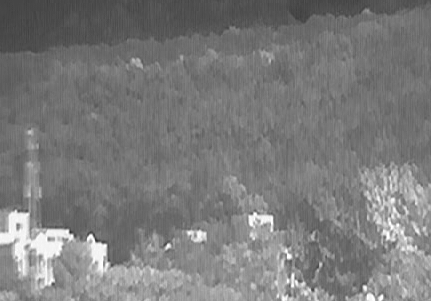} &
\includegraphics[height=1.9 cm]{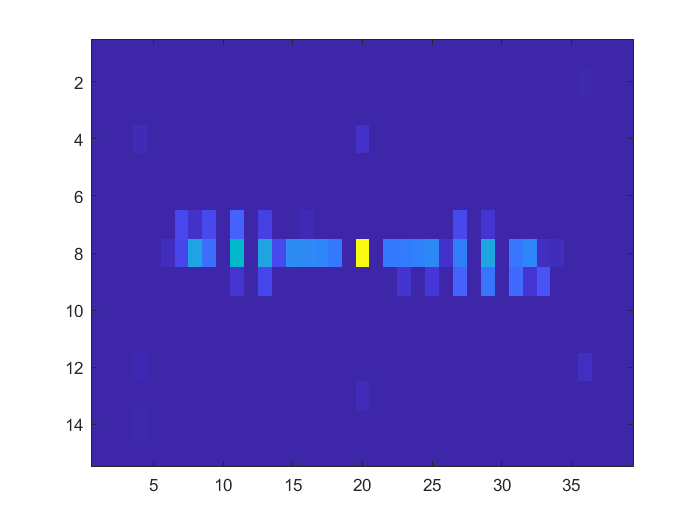} \\ 
(d) Mean of 9 frames & (e) Center-5th frame & (f) Kernel
\end{tabular}

\caption{Blur-sharp pairs and the estimated kernel obtained using Eq. \ref{blur_sharp_psf}. (a): steering rate 40 deg/sec, (d): steering rate 60 deg/sec.} 
\label{avg7} 
\end{figure}

Using Eq. (\ref{SR}), we determine the number of images \(N\) to be averaged for different steering rates (frame rate \(f_r=\) 30 frames/sec, exposure time \(t_{exp}=\) 5 ms). For steering rate 60 deg/sec: \(N\) is found to be 9; for steering rate 50 deg/sec: \(N=\) 8; for steering rate 40 deg/sec: \(N=\) 6; for steering rate 30 deg/sec: \(N=\) 5; for steering rate 20 deg/sec: \(N=\) 4; and for steering rate 10 deg/sec: \(N=\) 3. Examples of blur-sharp pairs formed from steering rate 1 deg/sec frames and the estimated blur kernels are given in Fig. \ref{avg7}. The PSFs thus estimated are next used in the non-blind deblurring methods to deblur the real IR images captured at corresponding steering rates. Note that the actual PSF in a real blurred image need not match the PSF estimated from blur-sharp pairs due to presence of noise. 

\begin{SCfigure*}
\scriptsize
\centering 
\setlength{\tabcolsep}{2pt}

\caption{Deblurred results with non-blind deblurring methods using the PSF estimated from blur-sharp pair. (a): steering rate 40 deg/sec, (e): steering rate 60 deg/sec.} 
\begin{tabular}{cccc}
\includegraphics[height=2.5 cm]{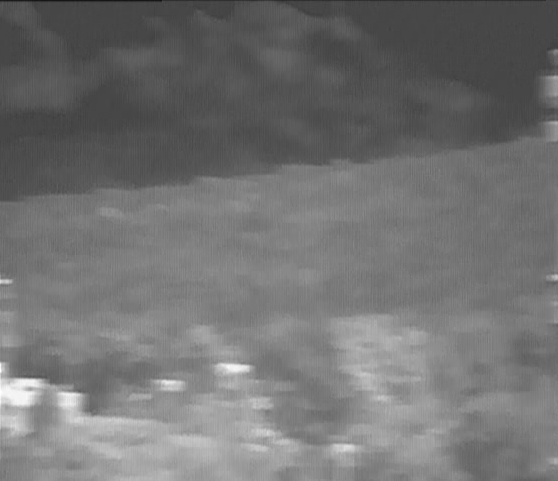} &
\includegraphics[height=2.5 cm]{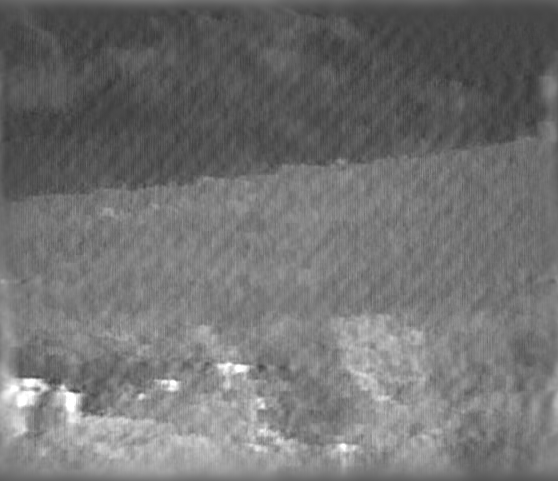} &
\includegraphics[height=2.5 cm]{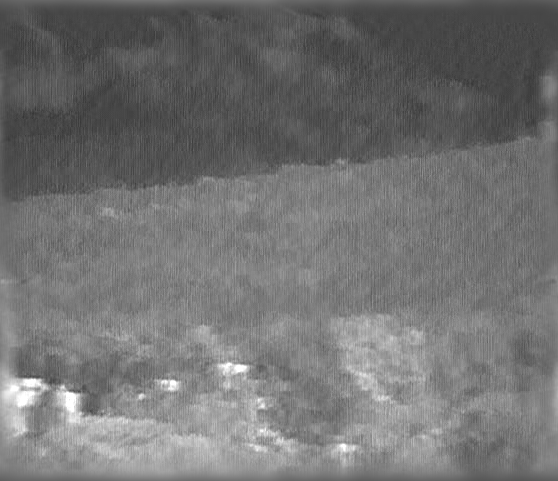}&
\includegraphics[height=2.5 cm]{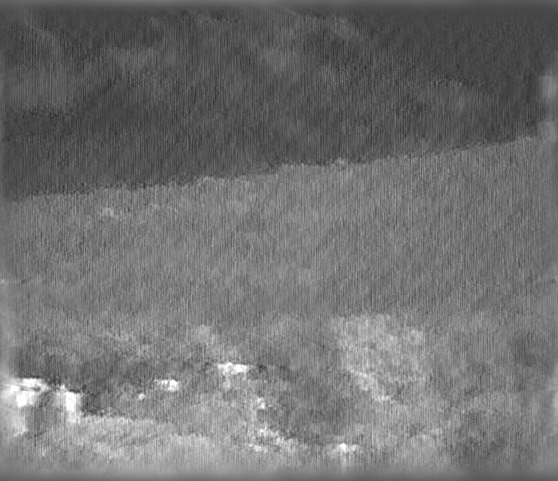}  \\ 
(a) Blurred - SR40 & (b) Krishnan et al. \cite{Ref5}&(c) RL  & (d) Wiener \\
\includegraphics[height=2.5 cm]{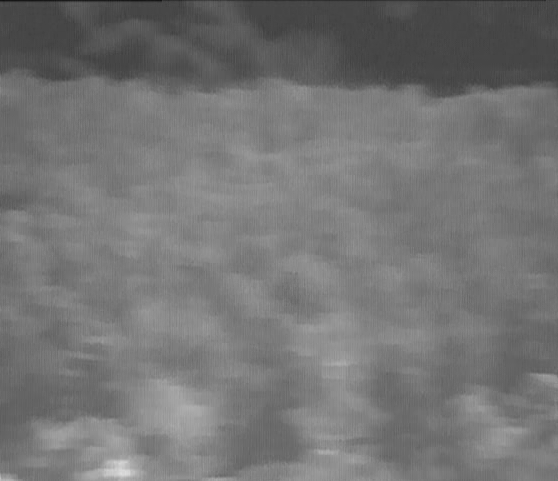} &
\includegraphics[height=2.5 cm]{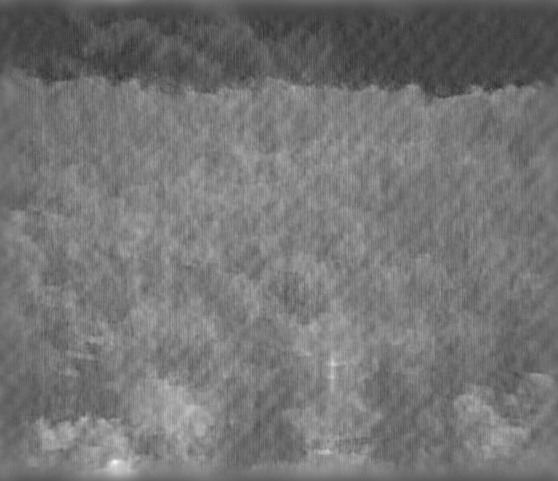} &
\includegraphics[height=2.5 cm]{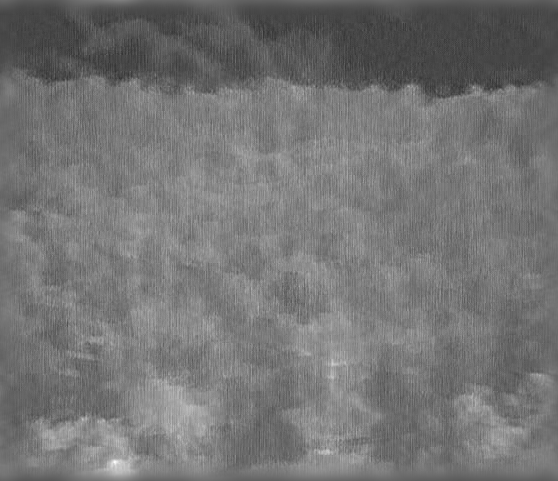}&
\includegraphics[height=2.5 cm]{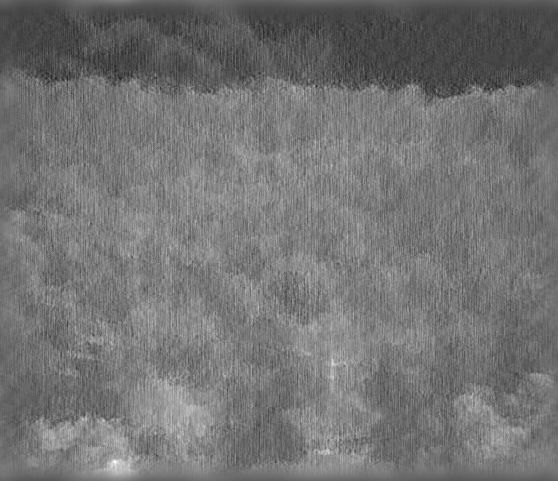}  \\ 
(e) Blurred - SR60 & (f) Krishnan et al. \cite{Ref5}&(g) RL &(h) Wiener 
\end{tabular}

\label{psfestimdeblur} 
\end{SCfigure*}
The deblurred results shown in Fig. \ref{psfestimdeblur} reveal good performance. Visually, we observe that the deblurring quality of Wiener filter is better than other methods. Krishnan et al. \cite{Ref5} has a few ringing artifacts.

\subsection{Results using analytical PSFs}
\label{analy_result}
To use the proposed method discussed in Section \ref{analy}, we need camera intrinsics. For the camera that we have used in our experiments, the FOV \((\alpha)=8\degree\) and the dimensions of the image are \(w=558\) pixels and \(h=481\) pixels. Thus \(f\) is determined using Eq. \ref{f}. Camera rotation angle \(\theta\) corresponding to one pixel spread upto the maximum pixel spread \(s\) is found using Eq. (\ref{eqn4}). The maximum pixel spread for each steering rate is selected using Eq. (\ref{s}) with \(t_{exp}\) = 5 ms.
\begin{figure}

\includegraphics[height=3.5 cm]{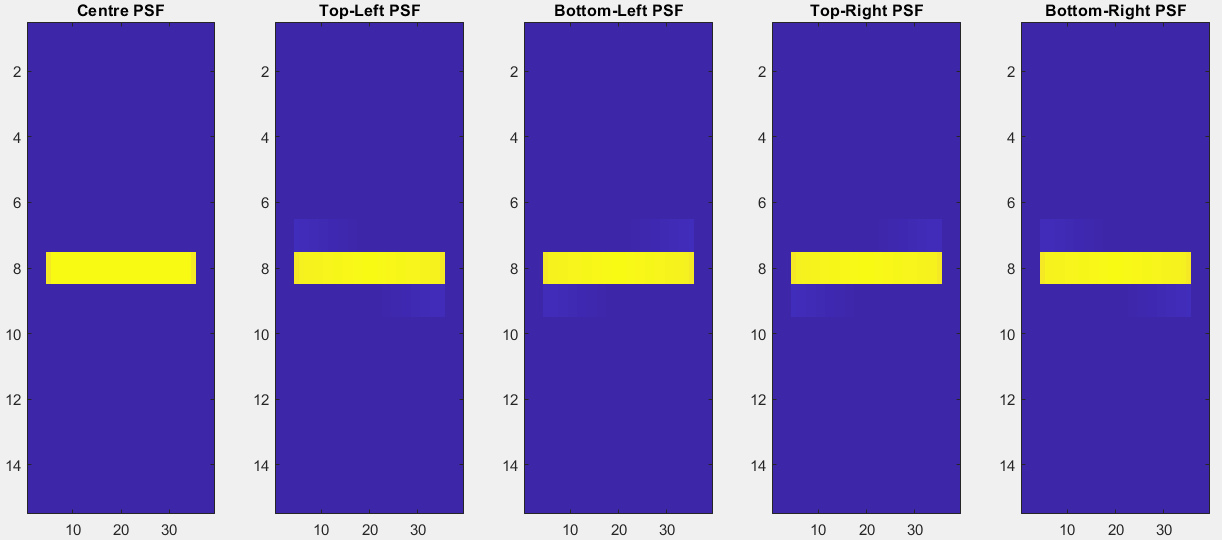}

\caption{Analytical PSFs obtained at different positions of the image.}
\label{fig_analy}       
\end{figure}
\begin{figure*}
\scriptsize
\centering 
\setlength{\tabcolsep}{2pt}
\begin{tabular}{cccc}

\includegraphics[height=2.8 cm]{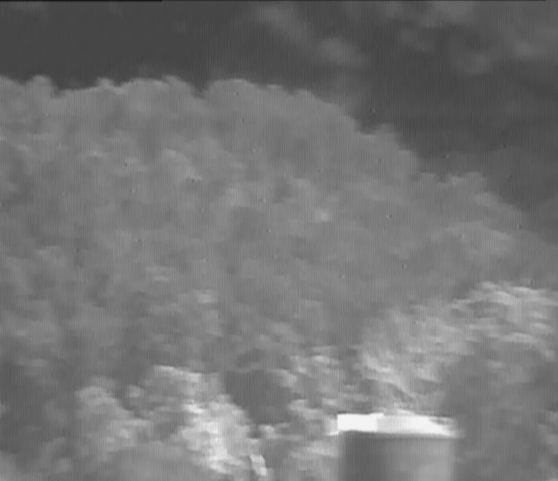} &
\includegraphics[height=2.8 cm]{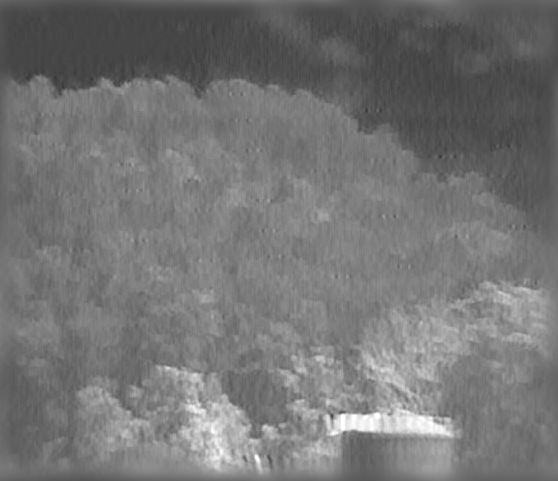}  &
\includegraphics[height=2.8 cm]{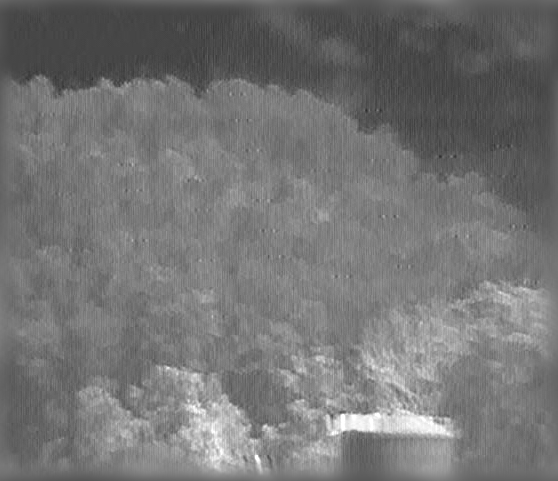}&
\includegraphics[height=2.8 cm]{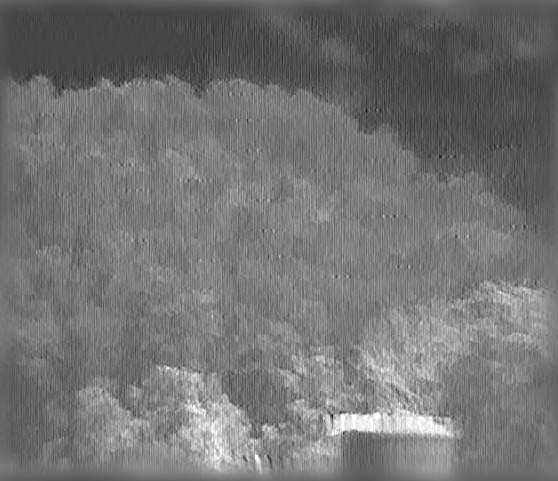} \\
(a) Blurred - SR40 & (b) Krishnan et al. \cite{Ref5}&(c) RL &(d) Wiener  \\

\includegraphics[height=2.8 cm]{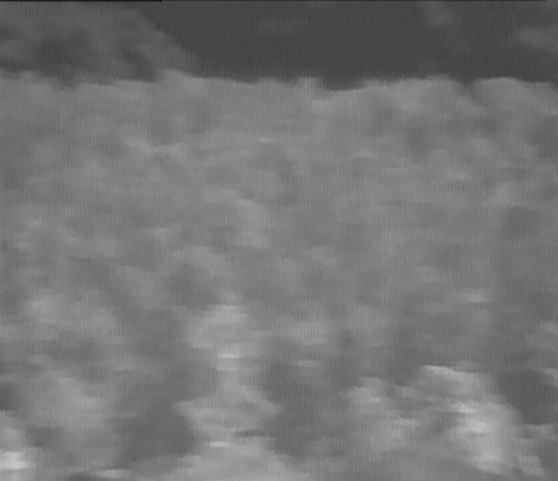} &
\includegraphics[height=2.8 cm]{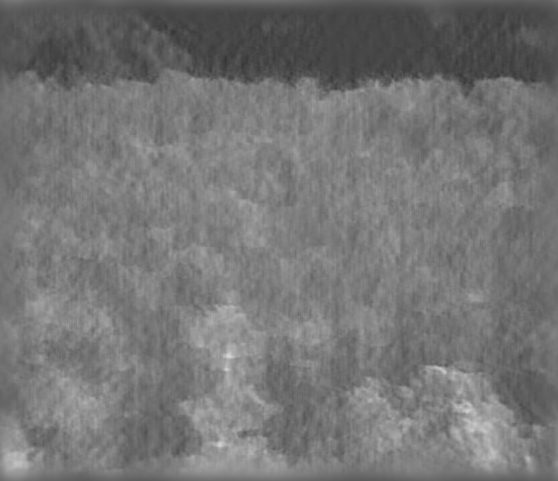} &
\includegraphics[height=2.8 cm]{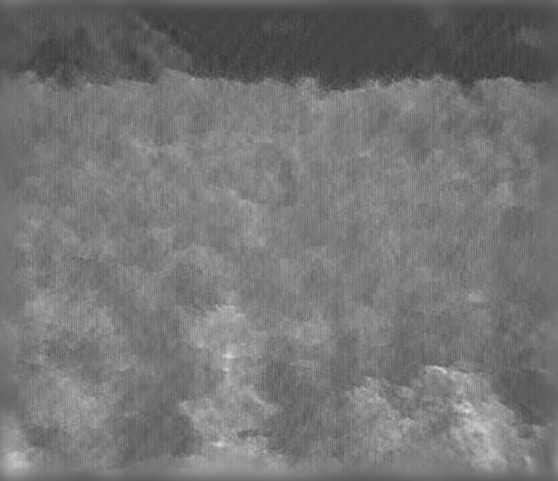}&
\includegraphics[height=2.8 cm]{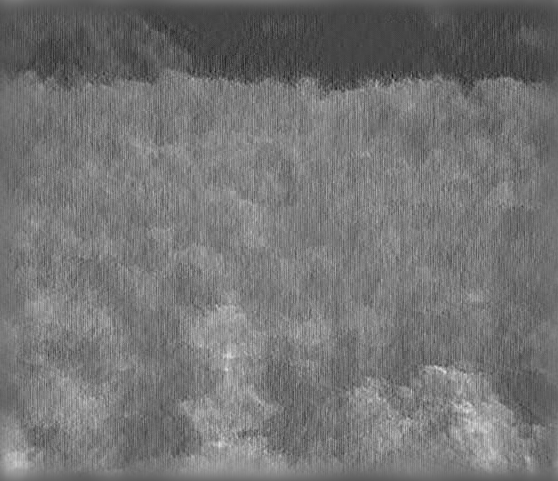}\\

(e) Blurred - SR60 & (f) Krishnan et al. \cite{Ref5}& (g) RL& (h) Wiener  
\end{tabular}

\caption{Deblurred results using analytical PSFs. (a): steering rate 40 deg/sec, (e): steering rate 60 deg/sec. Note that non-blind methods with analytical PSFs obtained using our proposed method exhibit quite good deblurring performance.} 
\label{analy_fig} 
\end{figure*}
By following the procedure described in Section \ref{analy}, the analytical PSFs obtained at different positions of the image  are given in Fig. \ref{fig_analy}. It shows that the blur is mostly space-invariant with some small variations at the edges of the kernels. We use the average of these 5 PSFs to deblur the real images and the corresponding deblurred results are given in Fig. \ref{analy_fig}. It can be observed that the deblurred results obtained using analytically modelled PSFs are quite good.

\begin{figure*}
\scriptsize
\centering 
\setlength{\tabcolsep}{2pt}
\begin{tabular}{cccc}
\includegraphics[height=2.8 cm]{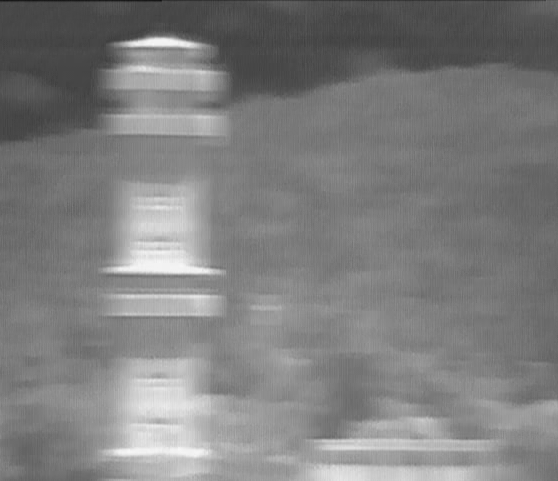} &
\includegraphics[height=2.8 cm]{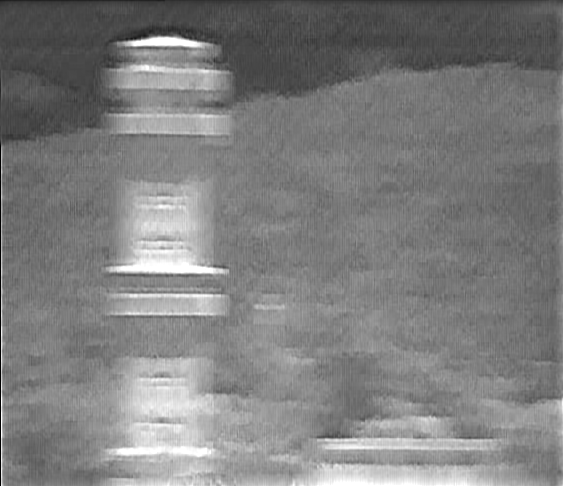} &
\includegraphics[height=2.8 cm]{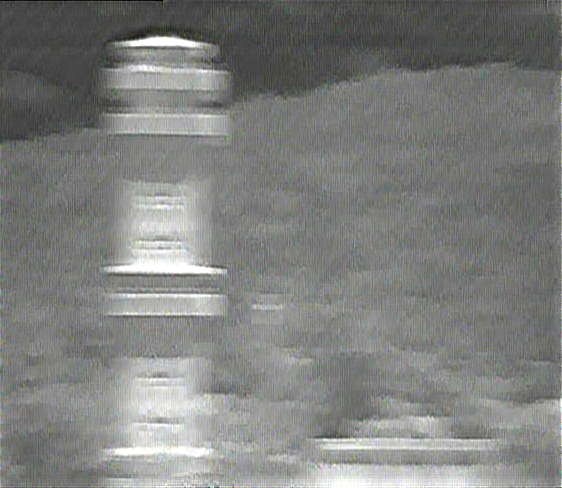} &
\includegraphics[height=2.8 cm]{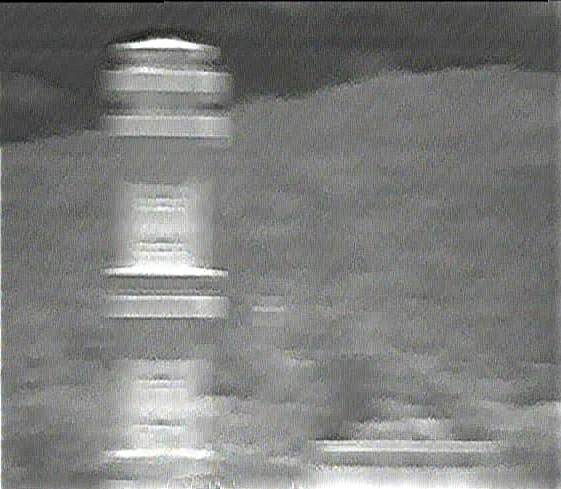}  \\
(A) Blurred& 1(a) & 2(a) & 3(a) \\

& \includegraphics[height=2.8 cm]{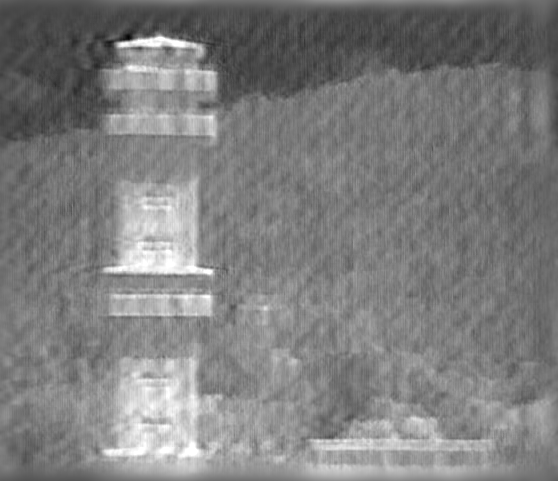} &
\includegraphics[height=2.8 cm]{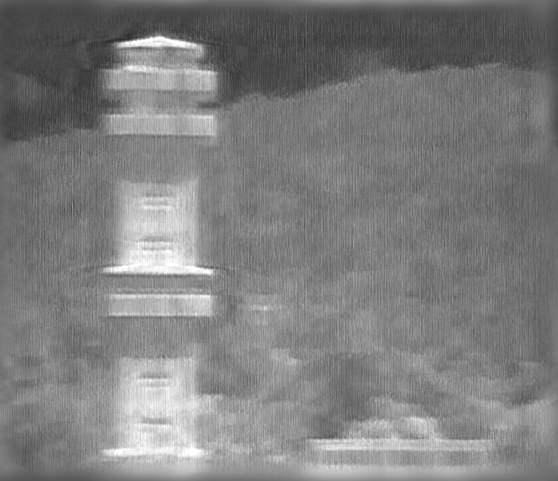} &
\includegraphics[height=2.8 cm]{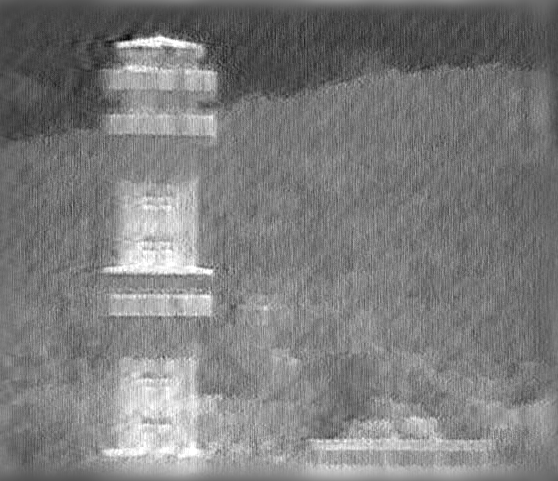} \\& 1(b) & 2(b) & 3(b)\\

& \includegraphics[height=2.8 cm]{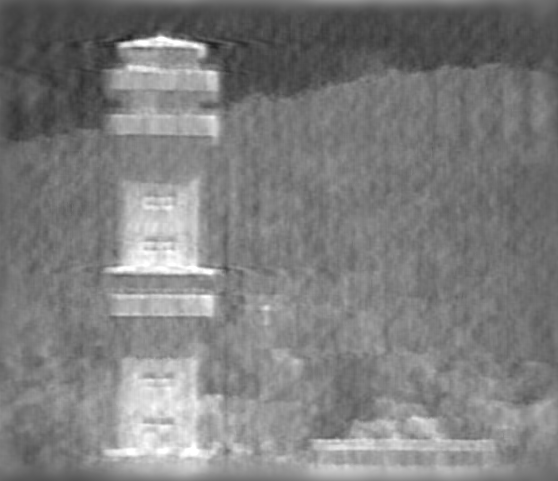}&
\includegraphics[height=2.8 cm]{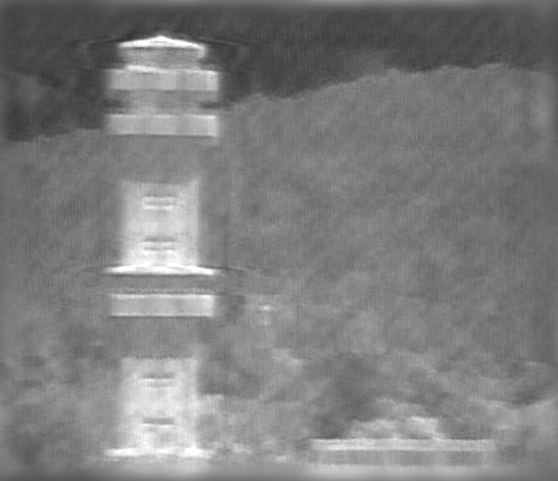}&
\includegraphics[height=2.8 cm]{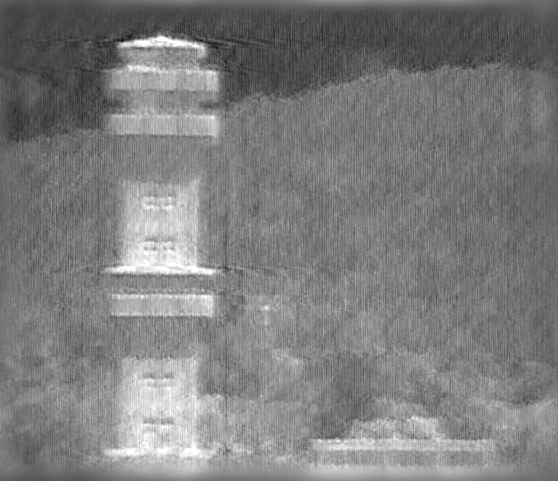}\\ & 1(c) & 2(c) & 3(c)\\

 & 1. Krishnan et al. \cite{Ref5}& 2. RL  & 3. Wiener \\
\end{tabular}
\caption{Comparisons of non-blind deblurring methods (1-3 in each column) for steering rate of 60 deg/sec using PSFs from three different methods (a: from blind method \cite{Ref11}, b: estimated using blur-sharp pair, c: analytical PSF). Note that our method (b) and (c) deliver the best performance consistently.}
\label{Compar_full} 
\end{figure*}

In Fig. \ref{Compar_full}, we give comparison results under one roof for a blurred image corresponding to steering rate of 60 deg/sec. It is amply evident that the results of our method (shown in b and c) yield best performance consistently across Wiener filter, RL as well as \cite{Ref5}. From the figure, it can be seen that the deblurred quality of non-blind methods using the estimated kernels from our approach is better than using the blur kernel returned from the blind deblurring method of \cite{Ref11}. Estimation of kernel using \cite{Ref11} is not only computationally expensive but also does not offer any advantage in getting good deblurring quality. Our PSF estimation methods are less computationally expensive and deliver good results. Visually, motion deblurring quality is better by using Krishnan et al. \cite{Ref5} and Wiener filter.

\subsection{Quantitative comparisons and Time complexity}
The nomenclature that we follow for the deblurring methods is as follows. 
\begin{enumerate}
    \item \(k_{blind}\): PSF obtained from the blind method of \cite{Ref11}.
    \item \(k_{BSpair}\): PSF estimated using blur-sharp pair.
    \item \(k_{anal}\): Analytical PSF derived using camera motion and intrinsic parameters.

\end{enumerate}

In order to quantitatively evaluate efficacy, from the steering rate of 1 deg/sec images, we synthetically generated 20 blurred images and added additive white Gaussian noise (AWGN) with signal-to-noise ratio (SNR) values of 32 dB, 35 dB and 38 dB. These images are then deblurred using the blind deblurring method of \cite{Ref11} as well as the non-blind deblurring methods (\(k_{blind}\), \(k_{BSpair}\) and \(k_{anal}\)) for all three SNR values. The average PSNR and SSIM values (higher is better) calculated over these 20 deblurred images are given in Table \ref{quality1}. Note that, in all the cases, our proposed method outperforms \cite{Ref11} as well as the non-blind deblurring methods with kernels derived from \cite{Ref11}. As the noise level increases, \cite{Ref11} deteriorates considerably due to kernel estimate errors. In contrast, our methods (especially \(k_{anal}\)) outperform the latter by a large margin in noisy scenarios.
 \begin{table*}
   \scriptsize
\caption{   Quantitative evaluations using PSNR and SSIM: NBD - Non-blind Deblurring.}
\label{quality1}       
    \centering
    \begin{tabular}{|m{1cm}|m{0.44cm}|p{0.44cm}|p{0.44cm}|p{0.43cm}|p{0.43cm}|p{0.43cm}|p{0.43cm}|p{0.43cm}|p{0.43cm}|p{0.43cm}|p{0.43cm}|p{0.43cm}|p{0.43cm}|p{0.43cm}|p{0.43cm}|p{0.43cm}|p{0.43cm}|p{0.43cm}|}
    \hline
     & \multicolumn{9}{c|}{\textbf{PSNR}(dB)$\uparrow$} & \multicolumn{9}{c|}{\textbf{SSIM}$\uparrow$} \\\hline
         \thead{Noise} & \multicolumn{3}{c|}{38 dB} & \multicolumn{3}{c|}{35 dB} & \multicolumn{3}{c|}{32 dB} & \multicolumn{3}{c|}{38 dB} & \multicolumn{3}{c|}{35 dB} & \multicolumn{3}{c|}{32 dB}  \\
        \hline
         \thead{\cite{Ref11}}& \multicolumn{3}{c|}{25.3} & \multicolumn{3}{c|}{25.2}& \multicolumn{3}{c|}{24.1}& \multicolumn{3}{c|}{0.63} & \multicolumn{3}{c|}{0.52}& \multicolumn{3}{c|}{0.37}\\
         \hline

         \textbf{NBD} &\thead{\cite{Ref5}} & \textbf{RL} & \textbf{WF} & \textbf{\cite{Ref5}} & \textbf{RL} & \textbf{WF}&\textbf{\cite{Ref5}} & \textbf{RL} & \textbf{WF} & \textbf{\cite{Ref5}} & \textbf{RL} & \textbf{WF}&\textbf{\cite{Ref5}} & \textbf{RL} & \textbf{WF} & \textbf{\cite{Ref5}} & \textbf{RL} & \textbf{WF} \\
         
         \hline
         \(k_{blind}\) & 27.6 & 26.2 & 26.5 & 27.1 & 24.9 & 26.2 & 25.9 & 23.1 & 25.2
         & 0.68 & 0.52 & 0.68 & 0.59 & 0.40 & 0.59 & 0.46 & 0.27 & 0.48\\
         \hline
         \(k_{BSpair}\) &  31.3 & 27.8 & 27.5 & 29.7 & 25.3 & 27.1 & 27.3 & 22.5
         & 25.9 & 0.76 & 0.54 & 0.70 & 0.66 & 0.38 & 0.61 & 0.50 & 0.26&0.49\\
         \hline
         \(k_{anal}\) &  29.9 & 29.2 & 27.9 & 29.0 & 28.1 & 27.5 & 27.6 & 26.2
         & 27.0 & 0.83 & 0.58 & 0.79 & 0.73 & 0.42 & 0.68 & 0.56 & 0.29&0.54\\
         \hline
    \end{tabular}
\end{table*}
\begin{table}
   \scriptsize
\caption{ Quantitative evaluations using no-Reference image quality scores.}
\label{quality}       
    \centering
    \begin{tabular}{||m{1.5cm}||c|c|c||c|c|c||}
    \hline
         \thead{Method} & \multicolumn{3}{c||}{\thead{NIQE$\downarrow$}} & \multicolumn{3}{c||}{\thead{PIQE$\downarrow$}}  \\
        \hline
         \thead{Blind\cite{Ref11}}& \multicolumn{3}{c||}{9.8} & \multicolumn{3}{c||}{47.5}\\
         \hline

         \thead{Non-blind \\Methods} &\thead{\cite{Ref5}} & \thead{RL} & \thead{Wie-\\ner} & \thead{\cite{Ref5}} & \thead{RL} & \thead{Wie-\\ner} \\
         
         \hline
         \(k_{blind}\) & 8.9 & 9.7 & 10.5 & 40.8 & 43.4 & 43.9\\
         \hline
         \(k_{BSpair}\) & 7.5 & 9.3 & 8.9 & 38.7 & 38.1 & 42.3\\
         \hline
         \(k_{anal}\) & 5.9 & 5.56 & 7.9 & 33.9 & 25.5 & 39.7\\
         \hline
    \end{tabular}
\end{table}

For the real deblurred images, since ground truth is not available, we use no-reference image quality scores such as naturalness image quality evaluator (NIQE) \cite{Refniqe} and perception based image quality evaluator (PIQE) \cite{Refpiqe} for quantitative evaluation. These scores are inversely correlated to the quality of image. The scores of different methods are given in Table \ref{quality}. The score of each method is its average score computed over 14 different deblurred images from steering rates 60 deg/sec and 50 deg/sec. It can be observed that non-blind deblurring methods again perform better than the state-of-the-art blind deblurring method of \cite{Ref11}. The quality of deblurred image using PSF estimated from our proposed method is better than using the PSF from the blind method. RL deconvolution output has less ringing and less noise which is reflected as lower scores (higher quality) in some cases. 

On an Intel Xeon e5-2630v4 processor \(@\) 2.2GHz, the blind deblurring method \cite{Ref11} processes a single frame in about 10 minutes. With a known PSF, and by using a single core of the processor, the processing time of the non-blind deblurring methods, Krishnan et al. \cite{Ref5}, RL and Wiener filter are 0.3 sec, 0.24 sec, and 0.04 sec, respectively. Even by using only a single core, the Wiener filter method delivers real-time performance (0.04 sec/frame).

  For further speed up, it is possible to use multiple cores to parallelise deblurring of several frames. Using 24 parallel workers, the processing time of different non-blind methods is given in Tables \ref{time2} and \ref{time3}. Table \ref{time2} gives the processing time for a real-time blurred video input (assuming 30 frames/sec) while Table \ref{time3} gives the processing time for real-time visualization of deblurred results of an already available blurred video (assuming 312 frames). Using parallel processing, Wiener filter achieves excellent real-time performance (0.014 sec/im-age) for a real-time video input. For the case of 312 frames, both the Wiener filter and Krishnan et al. \cite{Ref5} achieve real-time performance.

In summary, PSF estimated using our proposed methods followed by Wiener deconvolution is optimal both in terms of deblurring quality and real-time processing requirement. The PSF for each steering rate can be computed in advance using our proposed methods and can be stored in a Look Up Table (LUT). For a given steering rate of the input video, the corresponding PSF can be obtained from LUT and used in conjunction with the Wiener filter to get a real-time deblurred video output. 

\emph{Additional Remarks}: The ability to generate real blur-sharp pairs corresponding to a given steering rate has significant implication for the deep learning literature too. The availability of IR datasets for deblurring is quite scarce which precludes the use of deep learning methods. Importantly our approach can be used to generate a large number of blur-sharp pairs to emulate realistic motion blur in gimbal-based systems which can then be used to train deep networks. 
\begin{table}
   \scriptsize
\caption{Processing time for a real-time video input.}
\label{time2}     
    \centering
    
    \begin{tabular}{|c|c|c|}
    \hline
         \thead{Method} &  \multicolumn{2}{c|}{\thead{Execution time}}\\
         \cline{2-3}
          & \thead{per 30 frames} & \thead{per frame}\\
         \hline
         \thead{Krishnan et al. \cite{Ref5}}& 2.8 sec & 0.09 sec\\
         \hline
         \thead{RL deconv.}& 2.25 sec & 0.075 sec\\
         \hline
         \thead{Wiener deconv.}& 0.434 sec & \thead{0.014 sec}\\
         \hline
    \end{tabular}
\end{table}

\begin{table}
   \scriptsize
\caption{Processing time for real -time visualisation of an already available video.}
\label{time3}       
    \centering
 
    \begin{tabular}{|c|c|c|}
    \hline
         \thead{Method} & \multicolumn{2}{c|}{\thead{Execution time}}\\
         \cline{2-3}
          & \thead{per 312 frames} & \thead{per frame}\\
         \hline
         \thead{Krishnan et al. \cite{Ref5}}& 13.4 sec & \thead{0.04 sec}\\
         \hline
         \thead{RL deconv.}& 21.6 sec & 0.069 sec\\
         \hline
         \thead{Wiener deconv.}& 3.35 sec & \thead{0.0107 sec}\\
         \hline
    \end{tabular}
\end{table}

\section{Conclusions}
In this work, we dealt with the problem of real-time motion deblurring in gimbal-based systems deployed for surveillance and security applications. Our comprehensive studies on a real dataset captured from an actual gimbal-based system revealed that existing methods fall significantly short of delivering real-time deblurring performance. Towards this end, we model the motion blur either using a low steering rate dataset or analytically using the knowledge of camera motion and intrinsic parameters. In our method we use the estimated kernels in conjunction with non-blind deblurring methods to deliver excellent performance both in terms of deblurring quality and real-time processing. The blur-sharp pairs thus generated for creation of realistic datasets can be leveraged by deep learning networks too.

\end{document}